# Large Receptive Field Networks for High-Scale Image Super-Resolution

George Seif
Ryerson University
george.seif@ryerson.ca

Dimitrios Androutsos
Ryerson University
dimitri@ryerson.ca

## Abstract

*Convolutional Neural Networks have been the backbone of recent rapid progress in Single-Image Super-Resolution. However, existing networks are very deep with many network parameters, thus having a large memory footprint and being challenging to train. We propose Large Receptive Field Networks which strive to directly expand the receptive field of Super-Resolution networks without increasing depth or parameter count. In particular, we use two different methods to expand the network receptive field: 1-D separable kernels and atrous convolutions. We conduct considerable experiments to study the performance of various arrangement schemes of the 1-D separable kernels and atrous convolution in terms of accuracy (PSNR / SSIM), parameter count, and speed, while focusing on the more challenging high upscaling factors. Extensive benchmark evaluations demonstrate the effectiveness of our approach.*

## 1. Introduction

Single Image Super-Resolution (SISR) is a classic Computer Vision problem with the aim of recovering a High Resolution (HR) image from its Low Resolution (LR) version. It is a highly ill-posed problem because the LR image contains less data than the HR image and thus there are many possible mappings from the LR to HR space. In particular, high frequency details present in the Ground Truth (GT) HR image are often missing from the corresponding LR version, especially with large upscaling factors.

Recently, deep Convolutional Neural Networks (CNNs) have largely dominated the state-of-the-art in SISR beginning with the pioneering work of SRCNN [7]. Dong et al. trained their three-convolutional layer SRCNN to predict the ground-truth HR image from a bicubic upscaled LR image. In the image classification space Simonyan and Zisserman [22] have shown that increasing the overall receptive field of a deep CNN allows it to capture more spatial context and as a result achieve higher performance. Drawing from these ideas, further improvements were made to Super-Resolution (SR) networks by adding more layers [13] and using recursion [14] to increase network depth. The increase in depth results in a much larger receptive field and allows for learning of more complex image-to-image mappings. Due to the challenge of training very deep networks a number of works have proposed to use local skip connections via addition [17, 19, 24] or concatenation [28].

While deeper models can yield higher accuracy in terms of Peak Signal-to-Noise Ratio (PSNR) and Structural Similarity (SSIM) they also come with a number of drawbacks. Firstly, deeper models tend to have a very large number of parameters which leads to them requiring a larger memory bandwidth. Recursion has been successfully used in past works [24] to increase depth while mitigating memory consumption. However, in order to maintain state-of-the-art accuracy while minimizing memory consumption, a high number of recursions tend to be required and thus many layers. When using many layers a second drawback presents itself: some speed must be sacrificed for accuracy, even when using recursion to minimize memory consumption. Thirdly, due to both the large number of parameters and layers, deep networks are difficult to train, often requiring careful design of the learning rate schedule and gradient clipping.

In this paper, we discuss how to increase the receptive field of SR networks without increasing the number of layers or parameter count to have a more efficient use of parameters. In particular, we propose two different methods to expand the receptive field: atrous convolutions and one-dimensional separable filters. Atrous convolutions expand the receptive field by using spacing between the weights in the convolutional filters [31]. One-dimensional separable filters can expand the receptive field by use of large kernels. We show how both techniques can be applied to super-resolution networks to increase accuracy while maintaining layer and parameter count, without major sacrifices in speed. We explore several different separable filter and atrous convolution arrangement schemes. Our proposed Large Receptive Field Network (LRFNet) demonstrates state-of-the-art accuracy in terms of PSNR and SSIM, while being easy to train and requiring low memory.

The rest of this paper is structured as follows. Section 2 reviews the recent literature on Super-Resolution as well as other works in deep learning and computer vision that relate to our proposed methods. Section 3 describes our model explorations and proposed methods. In Section 4 we present our implementation details and experimental results on benchmark SR image datasets, including the NTIRE challenge validation dataset. We conclude our work in Section 5.

## 2. Related Work

### 2.1. Pre-Deep Learning SR

Early methods in SR were interpolation based [2, 18, 33], leveraging image edges to perform a structurly-guided interpolation. Those methods usually lacked the ability to produce HR images with high-frequency, realistic textures. To address this limitation, sparse coding based approaches use dictionairy learning to learn sparse signal representations for image patches[29, 30]. Neighborhood regression algorithms [26, 27] resconstruct HR images from a combination of HR image patches that have similar LR representations to the input and have also shown strong promise in solving the SISR problem.

### 2.2. Deep Convolutional Neural Networks for SR

Most recently, deep CNNs have dominated the state-of-the-art in SISR. Dong et al. [7] were the first to train a CNN to learn the mapping from a LR image to its corresponding HR ground-truth. LR image patches are upscaled to the HR image size using bicubic interpolation before being fed into SRCNN; this techinique is adopted in a number of works following SRCNN. Further improvements have been made by drawing ideas from the advances made in the image classification space. The design of Kim et al's Very Deep Super-Resolution (VDSR) network for SR [13] is inspired by the VGG network [22]. In particular, they stack 20 layers of 3x3 convolutions and Rectified-Linear Units (ReLUs) inbetween them to achieve a higher representational power than the smaller 3-layer SRCNN. They also introduce global residual learning with a skip connection summing the input and output of the network to address the vanishing/exploding gradients problem [4]. The Deeply-Recursive Convolutional Network (DRCN) [14] uses recursions to increase network depth while trading-off speed. However, even with a global residual, DRCN is difficult to train due to the exploding/vanishing gradients as a result of the high number of layers.

To address the strong tradeoff of sacrificing speed for higher performance, Dong et al. proposed the FSRCNN model [8] which directly processes the LR images without bicubic interpolation. To upscale the image to the HR size, a transposed convolution with strides equal to the SR scaling factor is utilized as the last layer of the network. Shi et al. use a similar technique with ESPCN [21], again processing at the lower-resolution, but using a sub-pixel convolution (fractional stride) to perform the upscaling. A Laplacian pyramid-like architecture is introduced in [16] as LapSRN which achieves both high performance and speed. LapSRN uses a stack of sub-networks where each sub-network processes images at their lower-resolution and upscales at the end similar to FSRCNN, while using a global residual like VDSR for each sub-network to ease training.

The most recent state-of-the-art networks have been very deep and use both local and global shortcut connections. Ledig et al. used a Generative Adversarial Network (GAN) [17] to tackle the SISR problem with SRResNet. They train a generator network to predict the HR image given the original LR image and additionally train a discriminator to distinguish between real HR images and those predicted by the generator. SRResNet uses local residuals inspired by ResNets [9, 10] to allieviate the vanishing/exploding gradient problem. Their network is very deep with 16 residual blocks. Tai et al. also use local residuals in their proposed Deep Recursive Residual Network (DRRN) [24], but drastically reduce the number of paramters compared to SRResNet by using recursion. Their DRRN achieves state-of-the-art accuracy while only having a single unique residual block and far less parameters than other models with comparable performance. The SRDenseNet model [28] has a similar structure to the SRResNet generator and LapSRN, processing the LR image before upscaling at the end. However, instead of using the local residual blocks from ResNets, they use the dense blocks from DenseNets [11] which have achieved higher performance in the image classification domain.

While these methods have shown strong promise in solving the SR problem they have a number of drawbacks. Expanding the network receptive field by adding more and more unique convolutional layers comes with the direct disadvantage of an increase in the number of parameters. The DRRN model addresses this by using recursive residual blocks, but when using less unique convolutional layers more recursions are required [24] resulting in a lower inference speed. As we will show, using atrous convolutions and one-dimensional separable kernels allows for lower parameter requirements while increasing performance in terms of PSNR and SSIM.

### 2.3. Related Works in Image Classification and Semantic Segmentation

Here we review a number of related works from the Image Classification and Semantic Segmentation domains that have inspired our proposed methods. Several state-of-the-art Image Classification algorithms have demonstrated that expanding the receptive field of CNNs, either by adding more layers [9, 22] or capturing multi-scale information [23] has a strong positive impact on the representational power and performance of deep CNNs. In the Semantic Segmentation space, Atrous convolutions have been used [6, 31] to expand the network receptive field and capture multi-scale context to achieve higher classification performance. We will show that atrous convolutions can be used directly for the task of SR without a pre-trained classification network frontend. Peng et al. proposed a Global Convolutional Network (GCN) [20] for Semantic Segmentation which uses large separable kernels and a boundary refinement module for an expanded receptive field while minimizing the number of parameters. We will show that separable kernels can also be applied to SR even without the use of a boundary refinement module with square kernels. All of these works demonstrate that a larger network receptive field can lead to higher performance.

## 3. Proposed Methods

In this section, we describe the technical design of our proposed Large Receptive Field Network (LRFNet). We explore the design space of SR networks using one-dimentional separable filters and atrous convolutions. To compare the performance of our models using one-dimensional separable filters and atrous convolutions to regular convolutions we first construct a baseline model as shown in Figure 1. The input to our model is a bicubic upscaled LR image. We use both global and local residual learning through global and local additive skip connections. The structure of each residual block is adopted from the EDSR model [19] and is shown in Figure 3a. We use 12 residual blocks and all convolutions in the network are 3x3 with 64 filters and ReLU non-linearities, except for the last one before the output which has no activation function. We denote our baseline model as LRFNet-B.

### 3.1. Expanding the Receptive Field of SR Networks

Expanding the receptive field of deep CNNs has been shown to increase performance on a variety of Computer Vision tasks [6, 20, 22, 31]. Having a larger receptive field allows the network to capture more spatial context; in the context of SISR, this increases the ability of the network to reconstruct larger and more complex edge structures.

Consider Figure 2a which shows the receptive fields of two succesive 3x3 convolutions. Each 3x3 convolution takes 9 parameters (excluding the bias) and thus with two successive 3x3s we achieve a receptive field of 5 in both the vertical and horizontal directions using a total of 18 parameters. However, we can achieve a larger receptive field with less parameters by using larger one-dimensional kernels. In Figure 2b we use two succesive 5x1 kernels. This expands the receptive field to a size of 9 in the vertical dimension yet uses less parameters with 10 than the previous case of the two successive 3x3 convolutions which used 18. As an example extension of this idea, one could use two 1x5 convolutions and two 5x1 convolutions to achieve a receptive field of 9 in both directions using only 20 parameters (10 parameters for each pair of one-dimensional convolutions). To achieve the same receptive field of 9 using only 3x3 convolutions (Figure 2a) we would need to use 4 successive 3x3's which has a total of 36 parameters. Increasing the network receptive field using only two-dimensional square kernels comes with the severe drawback of increasing the number of parameters by a large amount. The use of one-dimensional kernels gives us the ability to easily control the tradeoff between the network receptive field and the number of parameters without as much negative consequence.

The use of the one-dimensional kernels is inspired by separable filtering used in Image Processing. A separable filter is a convolutional filter that can be written as a matrix product of two other convolutional filters of lower dimensionality. This is typically a two-dimensional filter that can be separated into a product of two one-dimensional filters, thus reducing the computational load and number of parameters. One may question as to whether or not simply using only one-dimensional filters in a CNN will give the same performance as using square filters, since the one-dimensional filters may or may not turn out to be separable. We show in our experiments (Tables 2 and 3)that the network does in fact achieve greater performance using less parameters with one-dimensional filters due to an expanded receptive field, perhaps making the corrections to adjust for only using one-dimensional filters during training. We denote the model using these one-dimensional kernels, inspired by separable convolutions as LRFNet-S.

Atrous convolutions expand the network receptive field by using spacing between the weights in the convolutional filters [31]; this spacing is known as the dilation rate. Thus atrous convolutions have a wider view by looking at pixels that are further away from the center while using the same number of parameters as regular convolutions. A convolutional filter with a dilation rate of 1 has its weights separated by a distance of 1 and is equivalent to a regular convolutional filter; a convolutional filter with a dilation rate of 2 has its weights separated by a distance of 2 and so on. Figure 2c shows an expansion of the network receptive field using atrous convolutions. Two successive 3x3 convolutions are used but the second one has a dilation rate of 2. This expands the receptive field of the the two 3x3 convolutions from 5 in both dimensions to 7 in both dimensions all without increasing the number of parameters at all. We denote the model using atrous convolutions as LRFNet-A.

### 3.2. Exploring the Design Space of Separable Filters and Atrous Convolutions in SR Networks

To build all of our proposed models, the square convolutional filters in each residual block are swapped out for convolutions using one-dimensional separable kernels (LRFNet-S) or atrous convolutions (LRFNet-A). For the 1-D kernels, each ResBlock has one $1 \times k$ and one $k \times 1$ convolution. For the atrous convolutions we simply dilate the convolutions within certain ResBlocks.

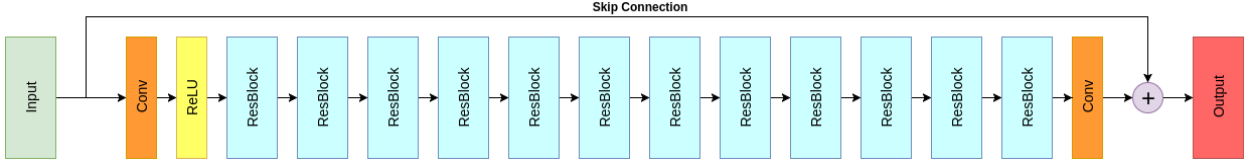

Figure 1: Our baseline model containing 12 residual blocks. Each residual block is composed of a conv-relu-conv structure and an additive skip connection.

Our baseline model uses the residual block structure shown in Figure 3a which has been shown to work well in SISR [19]. However, this structure was successfully tested using only square 3x3 kernels. Thus we conduct experiments using three other residual block schemes shown in Figure 3 with our one-dimensional separable filters. In Figure 3b we move the ReLU to be before the separable filters to test to see if allowing the feature maps to pass through the vertical and horizontal convolutions uninterrupted (i.e without the activation inbetween) improves performance. In Figures 3c and 3d we add the feature maps processed by the $1 \times k$ and the $k \times 1$ similar to the Global Convlutional Network (GCN) structure [20]. We test this to see if adding the feature maps from the $1 \times k$ and $k \times 1$ convolutions is more effective due independant processing in the vertical and horizontal directions. Activations before and after the convolutions are also tested. Note that we do not use any boundary refinement module as in GCN.

In addition to exploring different residual block schemes for one-dimensional kernels we also explore the design space for the atrous convolutions. The key questions we would like to answer are: How much dilation is required to increase performance? And does increasing the dilation rate always yield better performance? Specifically, we test out several settings of the dilation rate for the residual blocks. We denote the dilation rate as $\alpha$. Our baseline model has 12 residual blocks; among these residual blocks there are a total of 24 layers. We have tested the following dilation rate schemes: 1-2, 1-2-3, 1-3-5, and 1-4-8. A scheme with two dilation rates, for example 1-2, has the first 6 residual blocks with $\alpha = 1$ and the last 6 with $\alpha = 2$. A scheme with three dilation rates, for example 1-2-3, has the first 4 residual blocks with $\alpha = 1$, the second 4 with $\alpha = 2$, and the third with $\alpha = 3$.

Finally, we test to see if combining one-dimensional separable filters with dilation further enhances performance. We conduct this test using multiple 1-D kernel sizes combined with the single best performing dilation scheme.

## 4. Experiments

### 4.1. Implementation and Training Details

Our proposed models are built by swapping out the convolutions in each ResBlock for one-dimensional separable kernels or atrous convolutions. It is critical to note that all of our models have a total of 26 layers but vary in their total number of parameters. We train our models using the DIV2K dataset [25] which consists of 800 high-quality 2K resolution training images. To take full advantage of the larger network receptive field, we use patches of 128x128 pixels which is larger than most works (the same as LapSRN [16]), cropping the patches from the training data with no overlap. We use random rotation and flipping for data augmentation. The DIV2K dataset contains 100 2K resolution validation images which we use for evaluation of our model in the RGB colour space. We additionally evaluate our model on four well-known SISR benchmark datasets: Set5 [5], Set14 [32], BSDS100 [3], and Urban100 [12]. To facilitate fair comparison against the state-of-the-art on these benchmark datasets, we convert the output images to the YCbCr colour space and evaluate on the Y-Channel only. We record the PSNR, SSIM, inference time, and parameter count of all models.

We conduct initial experiments to test the effectiveness of one-dimensional kernels and atrous convolutions, as well as the various arrangement schemes, dilation rates, and kernel sizes. For these initial tests we use the first 100 images of the DIV2K dataset at $\times 4$ and $\times 8$ scale for training. As we will show, this is enough data to achieve results on par with state-of-the-art models and thus is sufficient for benchmarking of our proposed methods with the baseline. Following these experiments we select the best performing models and fully train them along with the baseline model using the entire DIV2K dataset of 800 images on both $\times 4$ and $\times 8$ scales. All scales are used to train the same model and thus both $\times 4$ and $\times 8$ SR is achieved using a single network. Training is done in Keras with TensorFlow backend [1] using the Adam optimizer [15]. The initial learning rate is set to $10^{-4}$ and divided by 2 every 50 epochs, training for a total of 300 epochs on a 1080Ti GPU.

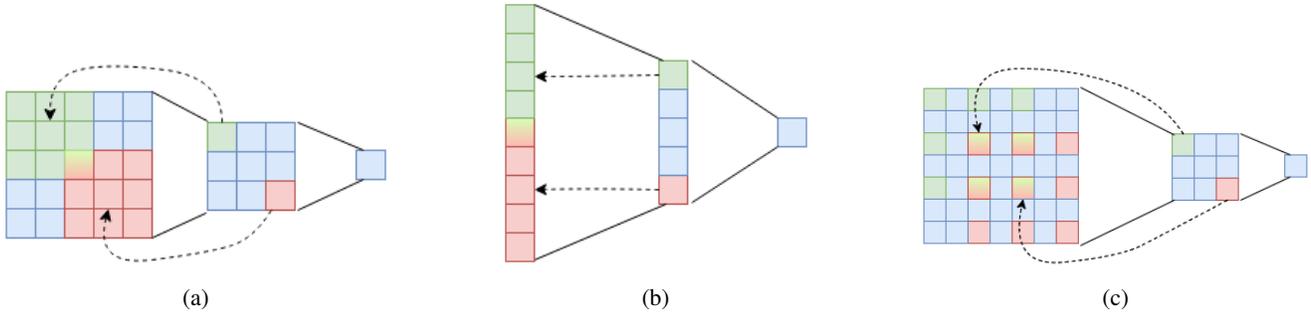

Figure 2: Network receptive field from two successive convolutions. (a) With two 3x3 convolutions we get an overall receptive field of 5 in both the vertical and horizontal directions. (b) Here we use two 5x1 convolutions and get an overall receptive field of 9 in the vertical direction, much larger in the vertical than using a square 3x3 but with less parameters (10 total vs 18 total in the 3x3 case). (c) Two successive 3x3 convolutions but the second convolution has a dilation rate of 2. This expands the overall receptive field from 5 in both directions (in (a)) to 7 in both directions without increasing the number of parameters.

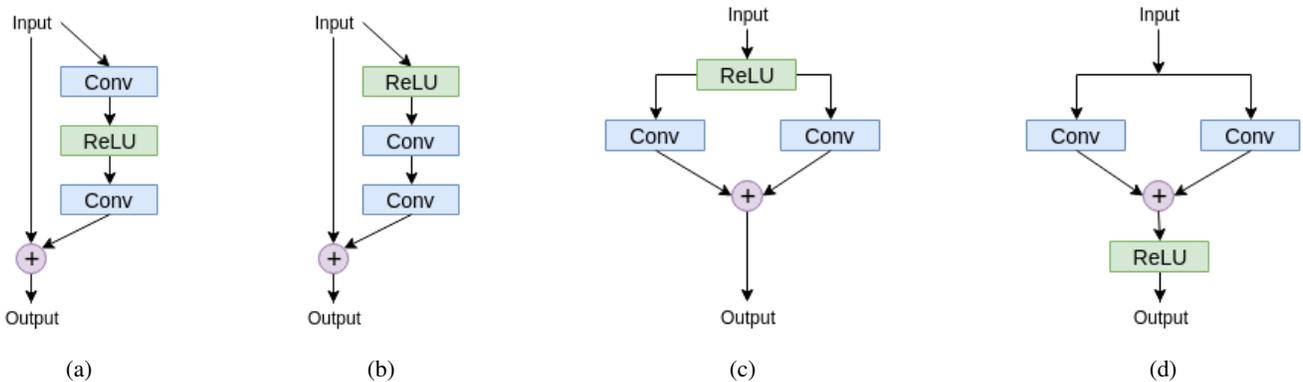

Figure 3: Exploring the network design space of local ResBlocks with separable filters. (a) Baseline residual block use in EDSR [19]. For the baseline model, we use this scheme with $k \times k$ convolutions. For our model with one-dimensional separable filters, the first convolution is $1 \times k$ and the second is $k \times 1$. (b) Moving the ReLU to before the separable filters. We test this to see if allowing the feature maps to pass through the vertical and horizontal convolutions uninterrupted (i.e without the activation inbetween) improves performance. (c) and (d) Adding the feature maps processed by the $1 \times k$ and the $k \times 1$. We test this to see if adding the feature maps from the $1 \times k$ and $k \times 1$ convolutions is more effective due independant processing in the vertical and horizontal directions. We test placing the activation before (c) and after (d) the convolutions.

## 4.2. Benchmarking LRFNet

### 4.2.1  1-D Separable Kernels: LRFNet-S

Here we benchmark the use of one-dimensional separable kernels in SISR. Our first experiment is determining whether the baseline ResBlock proposed for EDSR [19] is still optimal when using 1-D separable convolutions. We test out all of the schemes shown in Figure 3 using a kernel size of 7 where all kernels are one-dimensional. We denote them as schemes A to D for the Figures 3a to 3d. The results of our tests on the NTIRE validation set are shown in Table 1. The baseline ResBlock structure still has the best performance even when using one-dimensional separable kernels.

This ResBlock structure is used for the rest of our experiments.

We now demonstrate how expanding the network receptive field by increasing kernel size using one-dimensional separale kernels noticibly and consistently improves the performance of SISR models. Again the number of layers remains fixed in our experiments. We train our 1-D separable kernels model LRFNet-S using different kernel sizes and evaluate the model performance on the NTIRE validation set. We also do the same for the baseline model with square convolutions. Our results are shown in Table 2 for PSNR / SSIM and Table 3 for the Parameters / Inference Time.

| Scheme | Scale | PSNR / SSIM |
|---|---|---|
| A | ×4 | **28.32 / 0.804** |
|   | ×8 | **24.76 / 0.670** |
| B | ×4 | 28.28 / 0.803 |
|   | ×8 | 24.74 / 0.670 |
| C | ×4 | 28.27 / 0.802 |
|   | ×8 | 24.73 / 0.668 |
| D | ×4 | 28.27 / 0.803 |
|   | ×8 | 24.73 / 0.668 |

Table 1: ResBlock design for 1-D separable kernels. Tests are run on the NTIRE validation set.

| Kernel Size | Scale | Baseline PSNR / SSIM | LRFNet-S PSNR / SSIM |
|---|---|---|---|
| 3 | ×4 | 28.26 / 0.801 | 28.07 / 0.795 |
|   | ×8 | 24.69 / 0.668 | 24.54 / 0.662 |
| 5 | ×4 | 28.29 / 0.803 | 28.23 / 0.800 |
|   | ×8 | 24.74 / 0.670 | 24.69 / 0.668 |
| 7 | ×4 | **28.31 / 0.803** | 28.34 / 0.804 |
|   | ×8 | **24.75 / 0.670** | 24.76 / 0.671 |
| 9 | ×4 | 28.29 / 0.803 | 28.42 / 0.806 |
|   | ×8 | 24.72 / 0.670 | 24.77 / 0.671 |
| 11 | ×4 | 28.25 / 0.800 | **28.45 / 0.808** |
|   | ×8 | 24.69 / 0.668 | **24.79 / 0.672** |

Table 2: Comparing the performance of using different kernel size with 1-D separable kernels and the baseline. LRFNet-S indicates the model with 1-D separable kernels. Tests are run on the NTIRE validation set.

| Kernel Size | Baseline Params / Time | LRFNet-S Params / Time |
|---|---|---|
| 3 | 889k / 0.989 | 299k / 0.854 |
| 5 | 2,462k / 2.14 | 496k / 1.00 |
| 7 | 4,821k / 3.68 | 693k / 1.08 |
| 9 | 7,967k / 6.39 | 889k / 1.21 |
| 11 | 11,899k / 8.38 | 1,086k / 1.35 |

Table 3: Comparing the parameter count and inference time of using different kernel size with 1-D separable convolutions and the baseline. LRFNet-S indicates the model with 1-D separable kernels. Time is in seconds. Tests are run on the NTIRE validation set.

We can observe in Table 2 that increasing the size of 1-D separable kernels consistently improves performance in terms of PSNR and SSIM even up to a kernel size of 11. However, when using square convolutions with the baseline model, performance only increases up to a kernel size of 7 and then decreases for sizes 9 and 11. We can also see in Table 3 that when we increase kernel size for the baseline model the number of parameters and inference time rise rapidly; the consequence then of increasing kernel size for better performance is quite severe when using square kernels. When using 1-D separable kernels however we don't have as extreme of a tradeoff and we can exceed the performance of the model with square kernels using far less parameters. As an example, using kernel size 7, LRFNet-S outperforms LRFNet-B while using less than 15% of the parameters and running at over 3 times the inference speed. The results from both tables together show that as we increase the kernel size, the advantage of using 1-D separable kernels becomes more and more prominent in terms of performance (PSNR/SSIM), inference speed, and parameter count. For our final experiments with LRFNet-S, we set the kernel size to 9 to facilitate fair comparison to the baseline since both networks will have the same number of paramrters this way.

### 4.2.2 Atrous Convolutions: LRFNet-A

We now turn our attention to atrous convolutions for SR with LRFNet-A. The receptive field of atrous convolutions is controlled by the dilation rate where a higher dilation rate creates a larger receptive field. We conduct experiments to reveal how much dilation should be used and if increasing the dilation rate, which directly increases the receptive field, consistently improves performance. Results are shown in Table 4. Even with our most basic scheme of 1-2 using a dilation rate $\alpha = 2$ for the second set of 6 ResBlocks, we observe better performance both in terms of PSNR and SSIM compared to the baseline, still using the same number if parameters. As we introduce more dilation into the network with 3 different rates and as we increase each rate we observe consistently improving performance due to an expanded receptive field from schemes 1-2-3, to 1-3-5, to 1-4-8 all while using the same number of parameters as the baseline. We select the 1-4-8 scheme for our final LRF-A network due to its top performance on ×8 scale and strong performance on ×4 scale.

### 4.2.3 Can we get the Best of Both?: LRFNet-SA

In the previous sections it was shown that using large 1-D separable kernels and atrous convolutions independantly improves the performance of SISR networks. This naturally raises the question as to if we can get the best of both worlds by combining the two techniques i.e do dilated 1-D separable kernels lead to further performance improvements? We train models with 1-D separable kernels, each with a different kernel size but all using the same dilation scheme of 1-4-8 which had the best performance out of all those tested.

| Scheme | Scale | PSNR / SSIM | Time |
|---|---|---|---|
| Baseline | × 4 | 28.26 / 0.801 | 0.989 |
| | × 8 | 24.69 / 0.668 | |
| 1-2 | × 4 | 28.34 / 0.804 | 1.27 |
| | × 8 | 24.74 / 0.670 | |
| 1-2-3 | × 4 | 28.36 / 0.806 | 1.37 |
| | × 8 | 24.80 / 0.672 | |
| 1-3-5 | × 4 | **28.38 / 0.806** | 1.38 |
| | × 8 | 24.81 / 0.672 | |
| 1-4-8 | × 4 | 28.36 / 0.805 | 1.45 |
| | × 8 | **24.84 / 0.673** | |

Table 4: Comparing the performance of using different dilation schemes with Atrous convolutions all using 3x3 kernels. Time is in seconds. Tests are run on the NTIRE validation set.

| Kernel Size | Scale | LRFNet-SA PSNR / SSIM | LRFNet-S PSNR / SSIM |
|---|---|---|---|
| 3 | × 4 | 28.04 / 0.794 | 28.07 / 0.795 |
| | × 8 | 24.50 / 0.660 | 24.54 / 0.662 |
| 5 | × 4 | **28.16 / 0.798** | 28.23 / 0.800 |
| | × 8 | **24.62 / 0.665** | 24.69 / 0.668 |
| 7 | × 4 | 28.14 / 0.797 | 28.34 / 0.804 |
| | × 8 | 24.60 / 0.664 | 24.76 / 0.671 |
| 9 | × 4 | 28.10 / 0.796 | 28.42 / 0.806 |
| | × 8 | 24.58 / 0.663 | 24.77 / 0.671 |
| 11 | × 4 | 28.03 / 0.793 | **28.45 / 0.808** |
| | × 8 | 24.54 / 0.661 | **24.79 / 0.672** |

Table 5: Comparing the performance of using different 1-D separable kernel sizes with dilation. All models use the same dilation rate scheme of 1-4-8. Time is in seconds. Tests are run on the NTIRE validation set.

The results are shown in Table 5. As can be seen, combining dilation with 1-D separable kernels provides no benefit, in fact performing worse than just using 1-D separable kernels without dilation.

### 4.3. Comparison to the state-of-the-art

We compare our proposed methods to several state-of-the-art networks for SISR: SRCNN [7], VDSR [13], DRCN [14], LapSRN [16], DRRN [24]. For the ×8 scale, we use the quantitative results and images of Lai et al. [16] who retrained all of these state-of-the-art models for ×8 using the original settings from the papers. The quantitative results for these benchmark datasets are shown in Table 7. We also report our quantitative results for the DIV2K validation set in Table 6; in this case we evaluate our model in the RGB colour space.

| Algorithm | Scale | DIV2K Val. PSNR / SSIM |
|---|---|---|
| Baseline | × 4 | 28.63 / 0.812 |
| | × 8 | 25.03 / 0.679 |
| LRFNet-S | × 4 | 28.76 / 0.816 |
| | × 8 | 25.14 / 0.682 |
| LRFNet-A | × 4 | 28.68 / 0.814 |
| | × 8 | 25.24 / 0.684 |

Table 6: Comparing the performance of our fully trained models on both × 4 and × 8 scales for the NTIRE validation set. LRFNet-S uses 1-D kernels and LRFNet-A uses atrous convolutions

There are a number of key observations to make from our results. Expanding the receptive field using one-dimensional separable filters or atrous convolutions consistently improves SR performance, all while maintaining the number of parameters and depth of the network. For both ×4 and ×8 scales, our LRF networks achieve top performance by a large margin, while minimizing parameter count and depth, without the use of recursions. Due to being able to increase performance while maintaing depth, our models are also very easy to train and do not require gradient clipping.

An interesting observation is that our LRFNet-S performs better than LRFNet-A at ×4 scale, but LRFNet-A performs better at the ×8 scale. In fact, this can also be seen in our preliminary experimental results from Tables 2 and 4. This is likely due to the use of large dilation rates in LRFNet-A (the 1-4-8 scheme). Having such large gaps inbetween the convolution weights may be beneficial with ×8 scale, since the LR image itself for ×8 was obtained by skipping so many pixels. However, with ×4 scale, such large gaps may become detrimental and thus considering pixels that are adjacent to one another as in LRFNet-S helps to achieve better performance.

## 5. Conclusion

In this paper, we proposed two different Large Receptive Field Networks (LRFNets) for Single Image Super Resolution. The first network uses large 1-D seperable convolutional kernels (LRFNet-S) and the second uses atrous convolutions (LRFNet-A). Both expand the network receptive field from a standard baseline network that uses square kernel convolutions. We demonstrated that our LRFNets achieve state-of-the-art performance without increasing network depth or parameter count.

| Algorithm | Scale | Depth | Params. | Set5 | Set14 | BSDS100 | Urban100 |
|---|---|---|---|---|---|---|---|
| Bicubic | | - | - | 28.43 / 0.811 | 26.01 / 0.704 | 25.97 / 0.670 | 23.14 / 0.657 |
| SCRNN [7] | | 3 | 57k | 30.50 / 0.863 | 27.49 / 0.750 | 26.90 / 0.710 | 24.52 / 0.722 |
| VDSR [13] | | 20 | 665k | 31.35 / 0.883 | 28.01 / 0.767 | 27.29 / 0.725 | 25.18 / 0.752 |
| DRCN [14] | | 20 | 1775k | 31.54 / 0.884 | 28.03 / 0.768 | 27.24 / 0.725 | 25.14 / 0.752 |
| LapSRN [16] | ×4 | 24 | 812k | 31.54 / 0.885 | 28.19 / 0.772 | 27.32 / 0.727 | 25.21 / 0.756 |
| DRRN [24] | | 52 | 297k | 31.68 / 0.888 | 28.21 / 0.772 | 27.38 / 0.728 | 25.44 / 0.764 |
| Baseline | | 26 | 889k | 31.68 / 0.888 | 28.29 / 0.775 | 27.36 / 0.729 | 25.45 / 0.764 |
| LRFNet-S | | 26 | 889k | **31.91 / 0.890** | **28.44 / 0.778** | **27.47 / 0.733** | **25.70 / 0.773** |
| LRFNet-A | | 26 | 889k | 31.82 / 0.889 | 28.38 / 0.777 | 27.39 / 0.730 | 25.61 / 0.769 |
| Bicubic | | - | - | 24.40 / 0.658 | 23.10 / 0.566 | 23.67 / 0.548 | 20.74 / 0.516 |
| SCRNN [7] | | 3 | 57k | 25.33 / 0.690 | 23.76 / 0.591 | 24.13 / 0.566 | 21.29 / 0.544 |
| VDSR [13] | | 20 | 665k | 25.93 / 0.724 | 24.26 / 0.614 | 24.49 / 0.583 | 21.70 / 0.571 |
| DRCN [14] | | 20 | 1775k | 25.93 / 0.723 | 24.25 / 0.614 | 24.49 / 0.582 | 21.71 / 0.571 |
| LapSRN [16] | ×8 | 36 | 812k | 26.15 / 0.738 | 24.35 / 0.620 | 24.54 / 0.586 | 21.81 / 0.581 |
| DRRN [24] | | 52 | 297k | 26.18 / 0.738 | 24.42 / 0.622 | 24.59 / 0.587 | 21.88 / 0.583 |
| Baseline | | 26 | 889k | 26.46 / 0.753 | 24.46 / 0.626 | 24.63 / 0.589 | 21.98 / 0.590 |
| LRFNet-S | | 26 | 889k | 26.66 / 0.763 | 24.58 / 0.629 | 24.68 / 0.589 | 22.06 / **0.594** |
| LRFNet-A | | 26 | 889k | **26.77 / 0.765** | **24.68 / 0.631** | **24.73 / 0.590** | **22.09 / 0.594** |

Table 7: Test results on benchmark datasets. Red indicates the best performance and blue indicates the second best. Here, LRFNet-S uses a kernel size of 9 and LRFNet-A uses the 1-4-8 scheme.

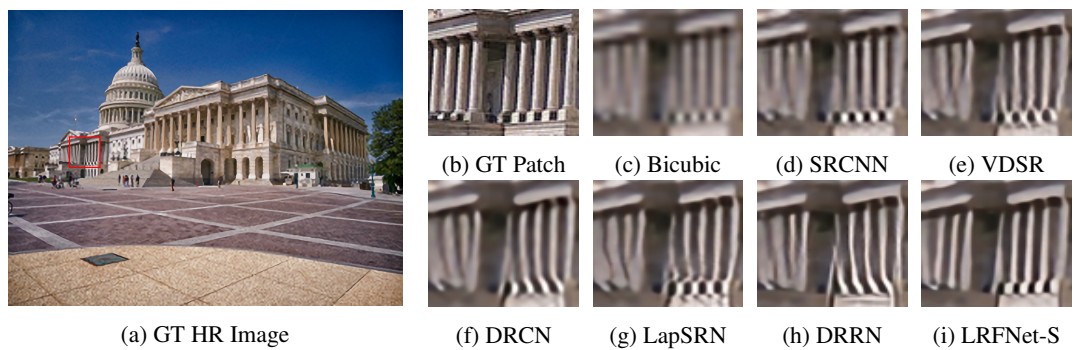

Figure 4: Qualitative comparison of our best performing model on ×4 scale with other works.

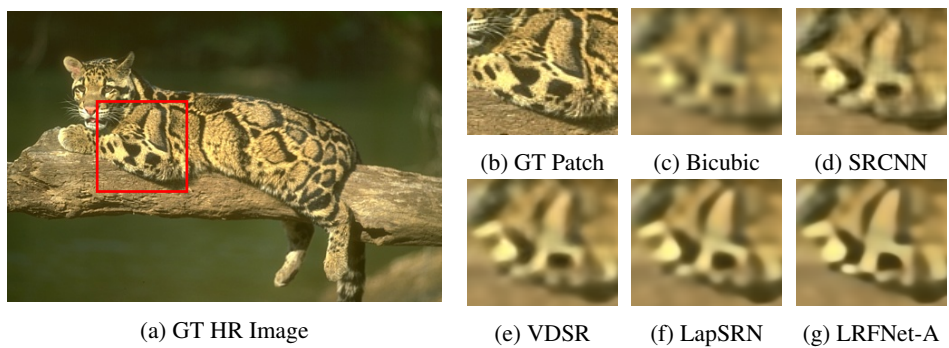

Figure 5: Qualitative comparison of our best performing model on ×8 scale with other works. Note that Lai et al's results for DRCN [14] and DRRN [24] were unavailable on their project page for ×8 scale and so are not shown here.